\documentclass[final]{cvpr}

\usepackage{times}
\usepackage{epsfig}
\usepackage{graphicx}
\usepackage{amsmath}
\usepackage{amssymb}

\usepackage{url}            %
\usepackage{booktabs}       %
\usepackage{amsfonts}       %
\usepackage{nicefrac}       %
\usepackage{microtype}      %
\usepackage{textcomp}
\usepackage[dvipsnames]{xcolor}
\usepackage{multirow}
\usepackage{flushend}
\usepackage{dsfont}

\DeclareMathSymbol{\shortminus}{\mathbin}{AMSa}{"39}

\usepackage{caption}
\usepackage{subcaption}
\usepackage{paralist}
\usepackage{array}
\usepackage{placeins}
\usepackage{epstopdf}
\usepackage[titletoc,title]{appendix}
\usepackage[T1]{fontenc}
\usepackage{diagbox}

\usepackage{transparent}

\usepackage{algorithm}
\usepackage{listings}

\usepackage{pifont}%

\usepackage{enumitem}
\setitemize{noitemsep,topsep=0pt,parsep=0pt,partopsep=0pt}

\newcommand{\parag}[1]{\smallskip\noindent\textbf{#1}~~}
\newcommand{\paragnoskip}[1]{\noindent\textbf{#1}~~}

\usepackage[pagebackref=true,breaklinks=true,colorlinks,citecolor=NavyBlue,bookmarks=false]{hyperref}

\def\cvprPaperID{7159} 
\def\confYear{CVPR 2021}
\usepackage{nopageno}

\newcommand{\vv}{\mathbf{v}}
\newcommand{\vw}{\mathbf{w}}
\newcommand{\vx}{\mathbf{x}}

\newcommand{\teacher}{\mathrm{T}}
\newcommand{\student}{\mathrm{S}}

\newcommand{\resnetfifty}{ResNet50\xspace}

\newcommand{\apbbox}[1]{AP$^\text{bb}_\text{#1}$}
\newcommand{\apmask}[1]{AP$^\text{mk}_\text{#1}$}

\makeatletter
\DeclareRobustCommand\onedot{\futurelet\@let@token\@onedot}
\def\@onedot{\ifx\@let@token.\else.\null\fi\xspace}

\def\etc{\emph{etc}\onedot} 
 
\def\etal{\emph{et al}\onedot}
\makeatother

\definecolor{codecomment}{rgb}{0.6,0.6,0.53}
\definecolor{codekeyword}{rgb}{0.65,0.1,0.1}
\definecolor{codeblue}{rgb}{0.25,0.5,0.5}
\definecolor{codegreen}{rgb}{0,0.6,0}
\definecolor{codegray}{rgb}{0.5,0.5,0.5}
\definecolor{codepurple}{rgb}{0.58,0,0.82}
\definecolor{backcolour}{rgb}{0.95,0.95,0.92}

\lstdefinestyle{mystyle}{
    backgroundcolor=\color{white},   
    commentstyle=\fontsize{7.2pt}{7.2pt}\color{codeblue},
    keywordstyle=\fontsize{7.2pt}{7.2pt}\color{codekeyword},
    numberstyle=\tiny\color{codegray},
    stringstyle=\color{codepurple},
    basicstyle=\fontsize{7.2pt}{7.2pt}\ttfamily\selectfont,
    columns=fullflexible,
    breakatwhitespace=false,         
    breaklines=true,                 
    captionpos=b,                    
    keepspaces=true,                 
    numbers=left,                    
    numbersep=5pt,                  
    showspaces=false,                
    showstringspaces=false,
    showtabs=false,                  
    tabsize=2
}

\begin{document}

\title{OBoW: Online Bag-of-Visual-Words Generation for Self-Supervised Learning}

\author{
Spyros Gidaris$^1$, \ \
Andrei Bursuc$^1$, \ \
Gilles Puy$^1$, \ \
Nikos Komodakis$^2$, \ \ 
Matthieu Cord$^{1,3}$,\ \
Patrick P\'erez$^1$\\
$^1$valeo.ai \ \ \ \ \ \ \  $^2$University of Crete\ \ \ \ \ \ \ $^3$Sorbonne Université\\
}

\maketitle

\begin{abstract}
Learning image representations without human supervision is an important and active research field. Several recent approaches have successfully leveraged the idea of making such a representation invariant under different types of perturbations, especially via contrastive-based instance discrimination training. Although effective visual representations should indeed exhibit such invariances, there are other important characteristics, such as encoding contextual reasoning skills, for which alternative reconstruction-based approaches might be better suited.

With this in mind, we propose a teacher-student scheme to learn representations by training a convolutional net 
to reconstruct a bag-of-visual-words (BoW) representation of an image, given as input a perturbed version of that same image. Our strategy performs an online training of both the teacher network (whose role is to generate the BoW targets) and the student network (whose role is to learn representations), 
along with an online update of the visual-words vocabulary (used for the BoW targets).
This idea effectively enables fully online BoW-guided unsupervised learning.
Extensive experiments demonstrate the interest of our BoW-based strategy, which surpasses previous state-of-the-art methods (including contrastive-based ones) in several applications.
For instance, in downstream tasks such Pascal object detection, Pascal classification and Places205 classification, our method improves over all prior unsupervised approaches, thus establishing new state-of-the-art results that are also significantly better even than those of supervised pre-training.
We provide the implementation code at {{\tt \small \url{https://github.com/valeoai/obow}}}.
\end{abstract}

\section{Introduction}

Learning unsupervised image representations based on convolutional neural nets (convnets) has attracted a significant amount of attention recently. Many different types of convnet-based methods have been proposed in this regard, including methods that rely on using annotation-free pretext tasks~\cite{doersch2015unsupervised,gidaris2018unsupervised,larsson2016learning,noroozi2016unsupervised,pathak2016context,zhang2016colorful}, generative methods that model the image data distribution \cite{donahue2017adversarial,donahue2019large,dumoulin2017adversarially}, as well as clustering-based approaches~\cite{asano2019selflabelling,caron2018deep,caron2019unsupervised}. 

Several recent methods opt to learn representations via instance-discrimination training~\cite{chen2020simple, dosovitskiy2014discriminative, he2020momentum, wu2018unsupervised}, typically implemented in a contrastive learning framework~\cite{chopra2005, hadsell2006dimensionality}.
The primary focus here is to learn low-dimensional image / instance embeddings that are invariant to intra-image variations while being discriminative among different images.
Although these methods manage to achieve impressive results, 
they focus less on other important aspects in representation learning, such as contextual reasoning, for which alternative reconstruction-based approaches~\cite{dapogny2020aaai, pathak2016context, zhang2016colorful, zhang2017split} might be better suited.
For instance, the task of predicting from an image region the contents of the entire image requires to recognize the visual concepts depicted in the provided region and then to infer from them the structure of the entire scene.
So, training for such a task has the potential of squeezing out more information from the training images and of learning richer and more powerful representations. 

However, reconstructing image pixels is an ambiguous and hard-to-optimize task that forces the convnet to spend a lot of capacity on modeling low-level pixel details. It is all the more unnecessary when the final goal deals with high-level image understanding, such as image classification.
To leverage the advantages of the reconstruction principle without focusing on unimportant pixel details, one can focus on reconstruction over high-level visual concepts, referred to as visual words. 
For instance, BoWNet~\cite{gidaris2020learning} derived a teacher-student learning scheme following this principle. 
In this teacher-student setting, given an image, the teacher extracts feature maps that are then quantized in a spatially-dense way over a vocabulary of visual words. Then, the resulting visual words-based image description is exploited for training the student on the self-supervised task of reconstructing the distribution of the visual words of an image, i.e., its bag-of-words (BoW) representation~\cite{yang2007evaluating}, given as input a perturbed version of that same image.
By solving this reconstruction task the student is forced to learn perturbation-invariant and context-aware representations while ``ignoring'' pixel details.

Despite its advantages,
the BoWNet approach exhibits some important limitations that do not allow it 
to fully exploit the potential of the BoW-based reconstruction task.
One of them is that it relies on the availability of an already pre-trained teacher network.
More importantly, it assumes that this teacher remains static throughout training.
However, due to the fact that during the training process the quality of the student's representations will surpass the teacher's ones, a static teacher is prone to offer a suboptimal supervisory signal to the student and to lead to an inefficient usage of the computational budget for training.

\begin{figure*}[t!]
\renewcommand{\captionfont}{\small}
\renewcommand{\captionlabelfont}{\bf}
\centering
\includegraphics[width=0.90\linewidth]{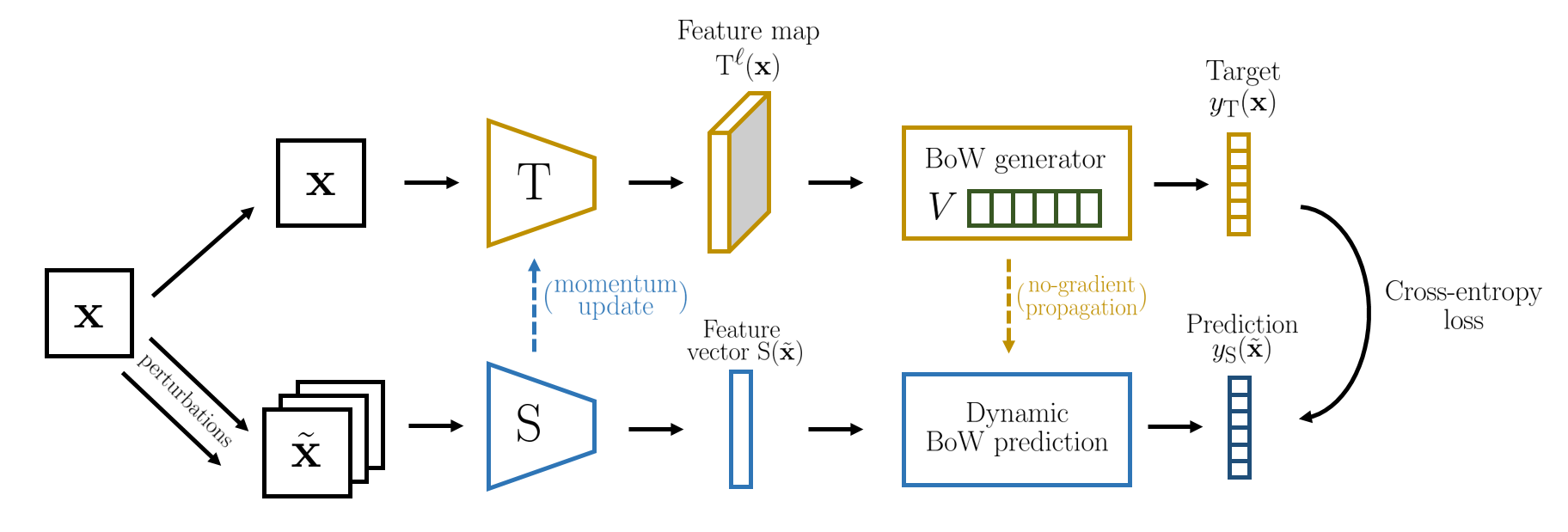}
\caption{\textbf{Unsupervised learning with Bag-of-Words guidance.} Two encoders $\teacher$ and $\student$ learn at different tempos by interacting and learning from each other. An image $\vx$ is passed through the encoder $\teacher$ and its output feature maps $\teacher^{\ell}(\vx)$ are embedded into a BoW representation $y_{\teacher}(\vx)$ over a vocabulary $V$ of features from $\teacher$. The vocabulary $V$ is updated at each step. The encoder $\student$ aims to reconstruct $y_{\teacher}(\vx)$ from data-augmented instances $\tilde{\vx}$. A dynamic BoW-prediction head learns to leverage the continuously updated vocabulary $V$ to compute the BoW representation from the features $\student(\tilde{\vx})$. $\teacher$ follows slowly the learning trajectory of $\student$ via momentum updates. }
\label{fig:pipeline}
\vspace{-12pt}
\end{figure*}

In this paper, we propose a BoW-based self-supervised approach that overcomes the aforementioned limitations. 
To that end, our main technical contributions are three-fold:
\begin{enumerate}[topsep=1ex,itemsep=0ex]
    \item 
    We design a novel \emph{fully online} teacher-student learning scheme for BoW-based self-supervised training (Fig.~\ref{fig:pipeline}). 
    This is achieved by online training of the teacher and the student networks. 
    \item We significantly revisit key elements of the BoW-guided reconstruction task.
    This includes the proposal of a \emph{dynamic} BoW prediction module used for reconstructing the BoW representation of the student image. This module is carefully combined with adequate online updates of the visual-words vocabulary used for the BoW targets.
    \item We enforce the learning of powerful contextual reasoning skills in our image representations. Revisiting data augmentation with aggressive cropping and spatial image perturbations, and exploiting \emph{multi-scale} BoW reconstruction targets, we equip our student network with a powerful feature representation. 
\end{enumerate}
Overall, the proposed method leads to a simpler and much more efficient training methodology for the BoW-guided reconstruction task that manages to learn significantly better image representations and therefore achieves or even surpasses the state of the art in several unsupervised learning benchmarks.
We call our method OBoW after the \textit{online BoW} generation mechanism that it uses.
\section{Related Work}

\paragnoskip{Bags of visual words.}
\emph{Bag-of-visual-words} representations are powerful image models able to encode image statistics from hundreds of local features.
Thanks to that, they were used extensively in the past~\cite{csurka2004visual,jegou2010aggregating,perronnin2007fisher,sivic2006video,tolias2013aggregate} and continue to be a key ingredient of several recent deep learning approaches~\cite{arandjelovic2016netvlad,gidaris2020learning,girdhar2017actionvlad,jain2020quest}.
Among them, BoWNet~\cite{gidaris2020learning} was the first work to use BoWs as reconstruction targets for unsupervised representation learning. 
Inspired by it, we propose a novel BoW-based self-supervised method with a more simple and effective training methodology that includes generating BoW targets in a fully online manner and further enforcing the learning of context-aware representations. 

\parag{Self-supervised learning.}
A prominent paradigm for unsupervised representation learning is to train a convnet on an artificially designed annotation-free pretext task,
e.g.,~\cite{arandjelovic2017look, chen2019self, doersch2015unsupervised, gidaris2018unsupervised, lee2017unsupervised,misra2016shuffle, noroozi2016unsupervised,vondrick2018tracking, wei2018learning, zhang2019aet, zhou2017unsupervised}.  
Many works rely on pretext reconstruction tasks
\cite{alayrac2019visual, godard2017unsupervised, larsson2016learning, pathak2016context, pillai2019superdepth, vincent2008extracting, zhang2016colorful, zhang2017split, zhou2017unsupervised}, 
where the reconstruction target is defined at image pixel level.
This is in stark contrast with our method, which uses a reconstruction task defined over high-level visual concepts (i.e., visual words) that are learnt with a teacher-student scheme in a fully online manner.

\parag{Instance discrimination and contrastive objectives.}
Recently, unsupervised methods based on contrastive learning objectives 
\cite{caron2020unsupervised,chen2020simple,falcon2020framework,frankle2020all,he2020momentum,henaff2020data,jaiswal2020survey,misra2020self,oord2018representation,tian2020contrastive,wang2020understanding,wu2018unsupervised} 
have shown great results. 
Among them, contrastive-based instance discrimination training \cite{chen2020simple,dosovitskiy2015discriminative,he2020momentum,kalantidis2020hard,wu2018unsupervised} is the most prominent example.
In this case, a convnet is trained to learn image representations that are invariant to several perturbations and at the same time discriminative among different images (instances). 
Our method also learns intra-image invariant representations, since the convnet must predict the same BoW target (computed from the original image) regardless of the applied perturbation.
More than that however, our work also places emphasis on learning context-aware representations, which, we believe, is another important characteristic that effective representations should have.
In that respect, it is closer to contrastive-based approaches that target to learn context-aware representations by predicting (in a contrastive way) the state of missing image patches~\cite{henaff2020data,oord2018representation,trinh2019selfie}.

\parag{Teacher-student approaches.} 
This paradigm has a long research history~\cite{gardner1989three, saad1996dynamics} and it is frequently used for distilling a single large network or an ensemble, the \emph{teacher}, into a smaller network, the \emph{student}~\cite{bucila2006model, hinton2015distilling, korattikara2015bayesian, papamakarios2015distilling}.
This setting has been revisited in the context of semi-supervised learning where the teacher is no longer fixed but evolves during training~\cite{laine2017temporal, tarvainen2017mean, verma2019interpolation}. In self-supervised learning, BowNet~\cite{gidaris2020learning} trains a student to match the BoW representations produced by a self-supervised pre-trained teacher. MoCo~\cite{he2020momentum} relies on a slow-moving momentum-updated teacher to generate up-to-date representations to fill a memory bank of negative images. BYOL~\cite{grill2020bootstrap}, which is a method concurrent to our work, also uses a momentum-updated teacher and trains the student to predict features generated by the teacher.
However, BYOL, similar to contrastive-based instance discrimination methods, uses low-dimensional global image embeddings as targets (produced from the final convnet output) and primarily focuses on making them intra-image invariant.
On the contrary, our training targets are produced by converting the intermediate teacher feature maps to high-dimensional BoW vectors that capture multiple \emph{local visual concepts}, thus constituting a richer target representation.
Moreover, they are built over an online-updated vocabulary from randomly sampled local features, expressing the current image as statistics over this vocabulary (see \S\,\ref{sec:full_online_training}). 
Therefore, our BoW targets expose fewer learning ``shortcuts'' (a critical aspect in self-supervised learning~\cite{arandjelovic2017look, doersch2015unsupervised, noroozi2016unsupervised, wei2018learning}), thus preventing to a larger extent teacher-student collapse and overfitting.

\parag{Relation to SwAV~\cite{caron2020unsupervised}.}
OBoW presents some similarity (e.g., using online vocabularies) with SwAV~\cite{caron2020unsupervised}.
However, the prediction tasks fundamentally differ: \textit{OBoW exploits a BoW prediction task while SwAV uses an image-cluster prediction task}
~\cite{asano2019selflabelling,caron2018deep,caron2019unsupervised}. BoW targets are much richer representations than image-cluster assignments: a BoW encodes all the local-feature statistics of an image whereas an image-cluster assignment encodes only one global image feature.

\section{Our approach}

Here we explain our proposed approach for learning image representations by reconstructing bags of visual words. 
We start with an overview of our method.

\parag{Overview.}
The bag-of-words reconstruction task involves a student convnet $\mathrm{S}(\cdot)$ that learns image representations, and a teacher convnet $\mathrm{T}(\cdot)$ that generates BoW targets used for training the student network. 
The student $\mathrm{S}(\cdot)$ is parameterized by $\theta_{\mathrm{S}}$ and the teacher $\mathrm{T}(\cdot)$ by $\theta_{\mathrm{T}}$.

To generate a BoW representation $y_{\mathrm{T}}(\vx)$ out of an image $\vx$, the teacher first extracts the feature map $\mathrm{T}^{\ell}(\vx) \in \mathbb{R}^{c_{\ell} \times h_{\ell} \times w_{\ell}}$, of spatial size $h_{\ell} \times w_{\ell}$ with $c_{\ell}$ channels, from its $\ell^{\rm th}$ layer (in our experiments $\ell$ is either the last $L$ or penultimate $L\!-\!1$ convolutional layer of $\mathrm{T}(\cdot)$).
It quantizes the $c_{\ell}$-dimensional feature vectors $\mathrm{T}^{\ell}(\vx)[u]$ at each location $u \in \{1, \cdots, h_{\ell} \times w_{\ell}\}$ of the feature map over a vocabulary $V = [\mathbf{v}_1, \ldots, \mathbf{v}_{K}]$ of $K$ visual words of dimension $c_{\ell}$.
This quantization process produces for each location $u$ a $K$-dimensional code vector $q(\vx)[u]$ that encodes the assignment of $\mathrm{T}(\vx)[u]$ to its closest (in terms of squared Euclidean distance) visual word(s).
Then, the teacher reduces the quantized feature maps $q(\vx)$ to a $K$-dimensional BoW, $\tilde{y}_{\mathrm{T}}(\vx)$, by channel-wise max-pooling, i.e., 
$\tilde{y}_{\mathrm{T}}(\vx)[k] = \max_{u} q(\vx)[u][k]$ (alternatively, the reduction can be performed with average pooling), where $q(\vx)[u][k]$ is the assignment value of the code $q(\vx)[u]$ for the $k^\mathrm{th}$ word.
Finally, $\tilde{y}_{\mathrm{T}}(\vx)$ is converted into a probability distribution over the visual words by $L_1$-normalization, i.e., 
$y_{\mathrm{T}}(\vx)[k] = \frac{\tilde{y}_{\mathrm{T}}(\vx)[k]}{\sum_{k'} \tilde{y}_{\mathrm{T}}(\vx)[k']}$.

To learn image representations, the student gets as input a perturbed version of the image $\vx$, denoted as $\tilde{\vx}$, and is trained to reconstruct the BoW representation $y_{\mathrm{T}}(\vx)$, produced by the teacher, of the original unperturbed image $\vx$.
To that end, it first extracts a global vector representation $\mathrm{S}(\tilde{\vx})\in \mathbb{R}^c$ (with $c$ channels) from the entire image $\tilde{\vx}$ and then applies a linear-plus-softmax layer to 
$\mathrm{S}(\tilde{\vx})$, 
as follows:
\begin{align}
y_{\mathrm{S}}(\tilde{\vx})[k] = \frac{ \exp( \vw_k^{\top} \mathrm{S}(\tilde{\vx}) )}{\sum_{k^\prime} \exp( \vw_{k^\prime}^{\top} \mathrm{S}(\tilde{\vx}))},
\label{eq:predictor}
\end{align}
where $W = [\vw_1, \cdots, \vw_K]$ are the $c$-dimensional weight vectors (one per word) of the linear layer.
The $K$-dimensional vector $y_{\mathrm{S}}(\tilde{\vx})$ is the predicted softmax probability of the target $y_{\mathrm{T}}(\vx)$. 
Hence, the training loss that is minimized for a single image $\vx$ is the cross-entropy loss
\begin{equation} \label{eq:bow_loss}
\mathrm{CE}\big(y_{\mathrm{S}}(\tilde{\vx}), y_{\mathrm{T}}(\vx)\big) = -\sum_{k=1}^K y_{\mathrm{T}}(\vx)[k] \log \big(y_{\mathrm{S}}(\tilde{\vx})[k]\big)
\end{equation}
between the softmax distribution $y_{\mathrm{S}}(\tilde{\vx})$ predicted by the student from the perturbed image $\tilde{\vx}$, and the BoW distribution $y_{\mathrm{T}}(\vx)$ of the unperturbed image $\vx$ given by the teacher.

\parag{Our technical contributions.}
In the following, we explain 
(i) in \S\,\ref{sec:full_online_training}, how to construct a fully online training methodology for the teacher, the student and the visual-words vocabulary, 
(ii) in \S\,\ref{sec:dynamic_bow_prediction}, how to implement a dynamic approach for the BoW prediction that can adapt to continuously-changing vocabularies of visual words, and finally 
(iii) in \S\,\ref{sec:contextual_reasoning_skills}, how to significantly enhance the learning of contextual reasoning skills by utilizing multi-scale BoW reconstruction targets and by revisiting the image augmentation schemes.

\subsection{Fully online BoW-based learning} \label{sec:full_online_training}

To make the BoW targets encode more high-level features, BoWNet pre-trains the teacher convnet $\mathrm{T}(\cdot)$ with another unsupervised method, such as RotNet~\cite{gidaris2018unsupervised}, and computes the vocabulary $V$ for quantizing the teacher feature maps off-line by applying k-means on a set of teacher feature maps extracted from training images.
After the end of the student training, during which the teacher's parameters remain frozen, the student becomes the new teacher $\mathrm{T}(\cdot) \gets \mathrm{S}(\cdot)$, a new vocabulary $V$ is learned off-line from the new teacher, and a new student is trained, starting a new training cycle.
In this case however, \textbf{(a)} the final success depends on the quality of the first pre-trained teacher, \textbf{(b)} the teacher and the BoW reconstruction targets $y_{\mathrm{T}}(\vx)$ remain frozen for long periods of time, which, as already explained, results in a suboptimal training signal, and \textbf{(c)} multiple training cycles are required, making the overall training time consuming. 

To address these important limitations, in this work we propose a fully online training methodology that allows the teacher to be continuously updated as the student training progresses, with no need for off-line k-means stages.
This requires an online updating scheme for the teacher as well as for the vocabulary of visual words used for generating the BoW targets, both of which are detailed below.

\parag{Updating the teacher network.} 
Inspired by MoCo~\cite{he2020momentum}, the parameters $\theta_{\mathrm{T}}$ of the teacher convnet are an exponential moving average of the student parameters. 
Specifically, at each training iteration the parameters $\theta_{\mathrm{T}}$ are updated as
\begin{equation} \label{eq:teacher_update}
\theta_{\mathrm{T}} \gets \alpha \cdot \theta_{\mathrm{T}} +  (1 - \alpha) \cdot \theta_{\mathrm{S}},
\end{equation}
where $\alpha \in [0, 1]$ is a momentum coefficient. 
Note that, as a consequence, the teacher has to share exactly the same architecture as the student.
With a proper tuning of $\alpha$, e.g., $\alpha=0.99$, this update rule allows slow and continuous updates of the teacher, avoiding rapid changes of its parameters, such as with $\alpha=0$,
which would make the training unstable. 
As in MoCo, for its batch-norm units, the teacher maintains different batch-norm statistics from the student. 

\parag{Updating the visual-words vocabulary.} 
Since the teacher is continuously updated, off-line learning of $V$ is not a viable option. 
Instead, we explore two solutions for computing $V,$ \emph{online k-means} and a \emph{queue-based vocabulary}.

\textbf{Online k-means.}
One possible choice for updating the vocabulary is to apply online k-means clustering after each training step. Specifically, as proposed in VQ-VAE~\cite{oord2017neural, razavi2019generating}, we use exponential moving average for vocabulary updates.
A critical issue that arises in this case is that, as training progresses, the features distribution changes over time. The visual words computed by online k-means do not adapt to this distribution shift leading to extremely unbalanced cluster assignments and even to assignments that collapse to a single cluster. In order to counter this effect, we investigate different strategies: (a) detection of rarely used visual words over several mini-batches and replacement of these words with a randomly sampled feature vector from the current mini-batch; (b) enforcing uniform assignments to each cluster thanks to the Sinkhorn optimization as in, e.g., \cite{asano2019selflabelling,caron2020unsupervised}. 
For more details see \S\ref{sec:appendix_online_kmeans}.

\begin{figure}[t!]
\renewcommand{\captionfont}{\small}
\renewcommand{\captionlabelfont}{\bf}
\centering
\includegraphics[width=0.85\linewidth]{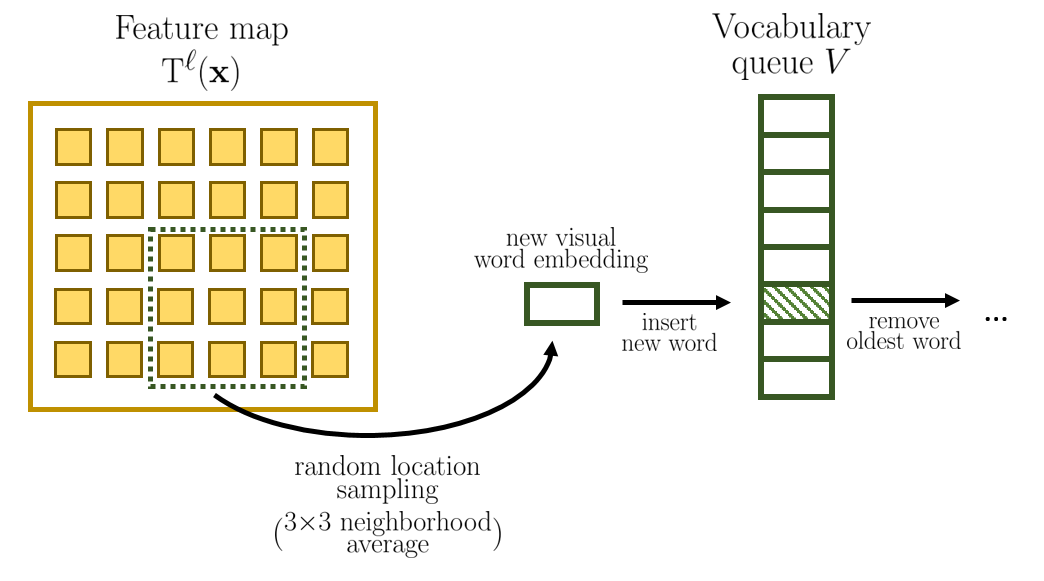}
\vspace{-10pt}
\caption{\textbf{Vocabulary queue from randomly sampled local features.} For each input image $\vx$ to  $\teacher$, ``local'' features are pooled from $\teacher^{\ell}(\vx)$ by averaging over $3 \times 3$ sliding windows. One of the resulting vectors is selected randomly and added as visual word to the vocabulary queue, replacing the oldest word in the vocabulary.}
\vspace{-10pt}
\label{fig:vocabulary_queue}
\end{figure}

\textbf{A queue-based vocabulary.} 
In this case, the vocabulary $V$ of visual words is a $K$-sized queue of random features.
At each step, after computing the assignment codes
over the current vocabulary $V,$ we update $V$ by selecting one feature vector per image from the current mini-batch, inserting it to the queue, and removing the oldest item in the queue if its size exceeds $K$. Hence, the visual words in $V$ are always feature vectors from past mini-batches. 
We explore three different ways to select these local feature vectors:
\textbf{(a)} uniform random sampling of one feature vector in $\mathrm{T}^{\ell}(\vx)$; 
\textbf{(b)} global average pooling of $\mathrm{T}^{\ell}(\vx)$ (average feature vector of each image);
\textbf{(c)} an intermediate approach between \textbf{(a)} and \textbf{(b)} which consists of a local average pooling with a $3 \times 3$ kernel (stride $1$, padding $0$) of the feature map $\mathrm{T}^{\ell}(\vx)$ followed by a uniform random sampling of one of the resulting feature vectors (Fig.\,\ref{fig:vocabulary_queue}).
Our intuition for option (c) is that, 
assuming that the local features in a $3 \times 3$ neighborhood belong to one common visual concept,
then local averaging selects a more representative visual-word feature from this neighborhood than simply sampling at random one local feature (option (a)).
Likewise, the global averaging option (b) produces a representative feature from an entire image, which however, might result in overly coarse visual word features.

The advantage of the queue-based solution over online k-means is that it is simpler to implement and it does not require any extra mechanism for avoiding unbalanced clusters, since at each step the queue is updated with new randomly sampled features. 
Indeed, in our experiments, the queue-based vocabulary with option \textbf{(c)} provided the best results.

\parag{Generating BoW targets with soft-assignment codes.} 
For generating the BoW targets, we use soft-assignments instead of the hard-assignments used in BoWNet. 
This is preferable from an optimization perspective due to the fact that the vocabulary of visual words is continuously evolving.
We thus compute the assignment codes $q(\vx)[u]$ as
\begin{equation}
q(\vx)[u][k] = \frac{\exp(-\frac{1}{\delta} \, \|\mathrm{T}^{\ell}(\vx)[u] - \mathbf{v}_k\|^2_2)}{\sum_{k^\prime} \exp(-\frac{1}{\delta} \, \|\mathrm{T}^{\ell}(\vx)[u] - \mathbf{v}_{k^\prime}\|^2_2)}.
\end{equation}
The parameter $\delta$ is a temperature value that controls the softness of the assignment.
We use $\delta = \delta_{\rm base} \cdot \bar{\mu}_{\text{MSD}}$, 
where $\delta_{\rm base} > 0$ and  $\bar{\mu}_{\text{MSD}}$ is the exponential moving average (with momentum $0.99$) of the mean squared distance of the feature vectors in $\mathrm{T}^{\ell}(\vx)$ from their closest visual words. 
The reason for using an adaptive temperature instead of a constant one is due to the change of magnitude of the feature activations during training, which induces a change of scale of the distances between the feature vectors and the words.

\begin{figure}[t!]
\renewcommand{\captionfont}{\small}
\renewcommand{\captionlabelfont}{\bf}
\centering
\includegraphics[width=0.80\linewidth]{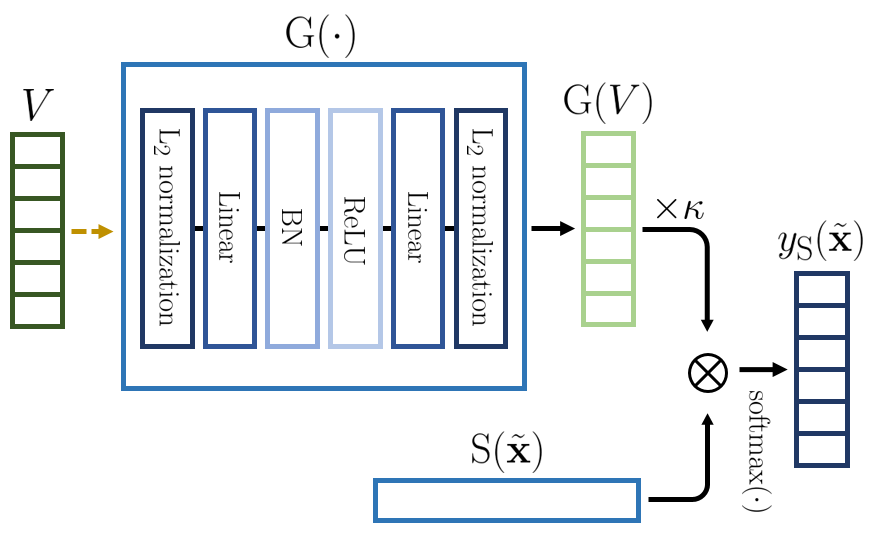}
\vspace{-10pt}
\caption{\textbf{Dynamic BoW-prediction head.} $\mathrm{G}(\cdot)$ learns to quickly adapt to the visual words in the continuously refreshed vocabulary $V$. The outputs $\mathrm{G}(V)$ are in fact weights that are used for mapping the features $\student(\tilde{\vx})$ to the corresponding BoW vector $y_{\student}(\tilde{\vx})$.
}
\vspace{-10pt}
\label{fig:bow_prediction_head}
\end{figure}

\subsection{Dynamic bag-of-visual-word prediction} \label{sec:dynamic_bow_prediction}

To learn effective image representations, the student must predict the BoW distribution over $V$ of an image using as input a perturbed version of that same image.
However, in our method the vocabulary is constantly updated and the visual words are changing or being replaced from one step to the next.
Therefore, predicting the BoW distribution over a continuously updating vocabulary $V$ with a fixed linear layer would make training unstable, if not impossible. 
To address this issue we propose to use a dynamic BoW-prediction head that can adapt to the evolving nature of the vocabulary.
To that end, instead of using fixed weights as in \eqref{eq:predictor}, we employ a generation network $\mathrm{G}(\cdot)$ that takes as input the current vocabulary of visual words $V = [\vv_1, \dots, \vv_K]$ and produces prediction weights for them as $\mathrm{G}(V) = [\mathrm{G}(\vv_1), \dots, \mathrm{G}(\vv_K)]$, where $\mathrm{G}(\cdot): \mathbb{R}^{c_{\ell}} \rightarrow \mathbb{R}^c$ is parameterized by $\theta_{\mathrm{G}}$ and $\mathrm{G}(\vv_k)$ represents the prediction weight vector for the $k^{\mathrm{th}}$ visual word.
Therefore, Equation\,\ref{eq:predictor} becomes
\begin{align}
y_{\mathrm{S}}(\tilde{\vx})[k] = \frac{\exp( \kappa \cdot \mathrm{G}(\vv_k)^{\top} \mathrm{S}(\tilde{\vx}) )}{\sum_{k^\prime} \exp( \kappa \cdot \mathrm{G}(\vv_{k^\prime})^{\top} \mathrm{S}(\tilde{\vx}))},
\label{eq:dynamic_predictor}
\end{align}
where $\kappa$ is a fixed coefficient that equally scales the magnitudes of all the predicted weights $\mathrm{G}(V) = [\mathrm{G}(\vv_1), \dots, \mathrm{G}(\vv_K)]$, which by design are $L_2$-normalized.
We implement $\mathrm{G}(\cdot)$ with a 2-layer perceptron whose input and output vectors are $L_2$-normalized (see Fig.\,\ref{fig:bow_prediction_head}).
Its hidden layer has size $2 \times c$.

We highlight that dynamic weight-generation modules are extensively used in the context of few-shot learning for producing classification weight vectors of novel classes using as input a limited set of training examples
\cite{gidaris2018dynamic,gomez2005evolving,qiao2018few}.
The advantages of using $\mathrm{G}(\cdot)$ instead of fixed weights, which BoWNet uses, are the equivariance to permutations of the visual words, the increased stability to the frequent and abrupt updates of the visual 
words, a number of parameters $|\theta_{\mathrm{G}}|$ independent from the number of visual words $K$, hence requiring fewer parameters than a fixed-weights linear layer for large vocabularies.

\subsection{Representation learning based on enhanced contextual reasoning} \label{sec:contextual_reasoning_skills}

\begin{figure}[t]
\renewcommand{\captionfont}{\small}
\renewcommand{\captionlabelfont}{\bf}
\begin{center}
\includegraphics[width=1.0\linewidth]{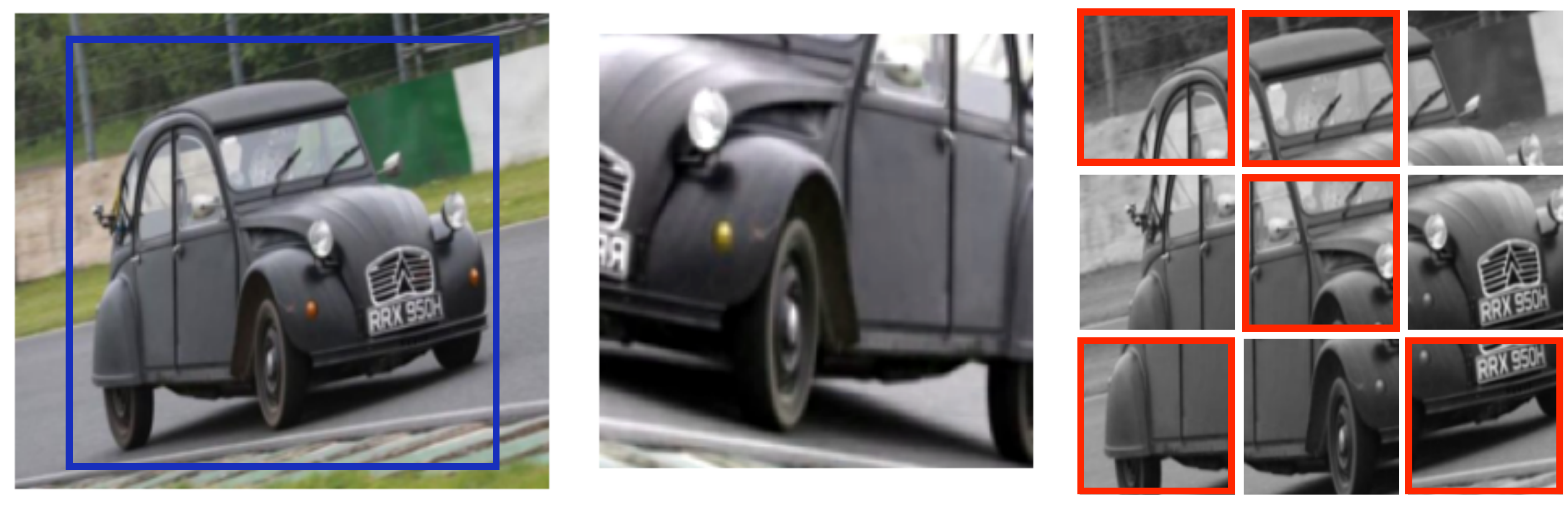}
\end{center}
\vspace{-20pt}
   \caption{
   \textbf{Reconstructing BoWs from small parts of the original image.}
   Given a training image (\textbf{left}), we extract two types of image crops.
   The first type (\textbf{middle}) is obtained by randomly sampling an image region whose area covers no more than $60\%$ of the entire image, resizing it to $160 \times 160$ and then giving it as input to the student as part of the reconstruction task. 
   The second type (\textbf{right}) is obtained by randomly selecting an area that covers between $60\%$ to $100\%$ of the entire image, resizing it to a $256 \times 256$ image, dividing it into $3 \times 3$ overlapping patches of size $96 \times 96$, and randomly choosing $5$ out of these $9$ patches (indicated with red rectangles) that are given as $5$ separate inputs to the student. The student must then reconstruct the original BoW target independently for each patch.
   The blue rectangle on the left image indicates the central $224 \times 224$ crop from which the teacher produces the BoW target.
   Note that, except from horizontal flipping, no other perturbation is applied on the teacher's inputs.
   }
   \vspace{-12pt}
\label{fig:img_aug}
\end{figure}

\parag{Data augmentation.}
The key factor for the success of many recent self-supervised representation learning methods~\cite{caron2020unsupervised,chen2020simple, chen2020improved, grill2020bootstrap,tian2020makes} is to leverage several image augmentation/perturbation techniques, such as Gaussian blur~\cite{chen2020simple}, color jittering and random cropping techniques, as cutmix~\cite{yun2019cutmix} that substitutes one random-size patch of an image with that of another.
In our method, we want to fully exploit the possibility of building strong image representations by hiding local information. 
As the teacher is randomly initialised, it is important to hide large regions of the original image from the student so as to prevent the student from relying only on low-level image statistics for reconstructing the distributions $y_{\mathrm{T}}(\mathbf{x})$ over the teacher visual words, which capture low-level visual cues at the beginning of the training.
Therefore, we carefully design our image perturbations scheme to make sure that the student has access to only a very small portion of the original image.
Specifically, similar to~\cite{caron2020unsupervised,misra2020self}, we extract from a training image multiple crops with two different mechanisms: one that outputs $160 \times 160$-sized crops that cover less than $60\%$ of the original image and one with $96 \times 96$-sized crops that cover less than $14\%$ of the original image (see Fig.\,\ref{fig:img_aug}). 
Given those image crops, the student must reconstruct the full bags of visual words from each of them independently.
Therefore, our cropping strategy definitively forces the student network to understand and learn spatial dependency between visual parts. 

\parag{Multi-scale BoW reconstruction targets.}
We also consider reconstructing BoW from multiple network layers that correspond to different scale levels. In particular, we experiment with using
both the last scale level $L$ (i.e., layer \texttt{conv5} in ResNet) and the penultimate scale level $L\!-\!1$ (i.e., layer \texttt{conv4} in ResNet).
The reasoning behind this is that the features of level $L\!-\!1$ still encode semantically important concepts but have a smaller receptive field than those in the last level.
As a result, the visual words of level $L\!-\!1$ that belong to image regions hidden to the student are less likely to be influenced by pixels of image regions 
the student is given as input.
Therefore, by using BoW from this extra feature level, the student is further enforced to learn contextual reasoning skills (and in fact, at a level with higher spatial details due to the higher resolution of level $L\!-\!1$), thus learning richer and more powerful representations. 
When using BoW extracted from two layers, our method includes a separate vocabulary for each layer, denoted by $V_{L}$ and $V_{L-1}$ for layers $L$ and $L\!-\!1$ respectively, and two different weight generators, denoted by $\mathrm{G}_{L}(\cdot)$ and $\mathrm{G}_{L\!-\!1}(\cdot)$ for layers $L$ and $L\!-\!1$, respectively.
Regardless of what layer the BoW target comes from, the student uses a single global image representation $\mathrm{S}(\tilde{\vx})$, typically coming from the global average pooling layer after the last convolutional layer (i.e., layer \texttt{pool5} in ResNet), to perform the reconstruction task.

We show empirically in Section \ref{sec:method_analysis} that the contextual reasoning skills implicitly developed via using the above two schemes are decisive to learn effective image representations with the BoW-reconstruction task.
\section{Experiments and results}

We evaluate our method (OBoW) on the ImageNet~\cite{russakovsky2015imagenet}, Places205~\cite{NIPS2014_5349} and VOC07~\cite{Everingham10} classification datasets as well as on the V0C07+12 detection dataset. 

\parag{Implementation details.}
For our models, the vocabulary size is set to $K=8192$ words and, as in BoWNet, when computing the BoW targets we ignore the visual words that correspond to the feature vectors on the edge of the teacher feature maps.
The momentum coefficient $\alpha$ for the teacher updates is initialized at $0.99$ and is annealed to $1.0$ during training with a cosine schedule.
The hyper-parameters $\kappa$ and $\delta_{\rm base}$ are set to $5$ and $1/10$ respectively for the results in \S\,\ref{sec:method_analysis}, to $8$ and $1/15$ respectively for the results in \S\,\ref{sec:large_scale_experiments}.
For more implementation details see \S\,\ref{sec:appendix_implementation_details}.

\subsection{Analysis} \label{sec:method_analysis}

Here we perform a detailed analysis of our method.
Due to the computationally intensive nature of pre-training on ImageNet, 
we use a smaller but still representative version created by keeping only $20\%$ of its images
and we implement our model with the light-weight ResNet18 architecture.
For training we use SGD for $80$ epochs with cosine learning rate initialized at $0.05$, batch size $128$ and weight decay $5\mathrm{e-}4$.
We evaluate models trained with two versions of our method, the vanilla version that uses single-scale BoWs and from each training image extracts one $160 \times 160$-sized crop (with which it trains the student), and the full version that uses multi-scale BoWs and extracts from each training image two $160 \times 160$-sized crops plus five $96 \times 96$-sized patches.

\parag{Evaluation protocols.}
After pre-training, we freeze the learned representations and use two evaluation protocols.
\textbf{(1)} The first one consists in training $1000$-way linear classifiers for the ImageNet classification task.
\textbf{(2)} 
For the second protocol, our goal is to analyze the ability of the representations to learn with few training examples. To that end, we use $300$ ImageNet classes and run with them multiple ($200$) episodes of $50$-way classification tasks with $1$ or $5$ training examples per class and a Prototypical-Networks~\cite{snell2017prototypical} classifier.

\subsection{Results}

\begin{table}[t!]
\centering
\renewcommand{\figurename}{Table}
\renewcommand{\captionlabelfont}{\bf}
\renewcommand{\captionfont}{\small} 
{\setlength{\extrarowheight}{1pt}\small
{
\begin{tabular}{ l  | c   c  | c  }
\toprule
& \multicolumn{2}{c|}{Few-shot} & \multicolumn{1}{l}{}\\
\multicolumn{1}{l|}{Updating method} & \multicolumn{1}{c}{1-shot} & \multicolumn{1}{c|}{5-shot} & \multicolumn{1}{l}{Linear}\\
\midrule
\multicolumn{4}{l}{\textbf{Online k-means}} \\
\;(a) replacing rare clusters   & 40.98 & 60.35 & 44.45\\
\;(b) Sinkhorn-based balancing  & 37.20 & 55.22 & 39.74\\
\midrule
\multicolumn{4}{l}{\textbf{Queue-based vocabulary}} \\
\;(a) local features             & 40.29 & 60.81 & 44.39\\
\;(b) globally-averaged features & 41.57 & \textbf{62.54} & 45.79\\
\;(c) locally-averaged features  & \textbf{42.11} & 62.44 & \textbf{45.86}\\
\midrule
\multicolumn{4}{l}{\textbf{Queue-based vocabulary -- multi-scale BoW}} \\
\;(b) globally-averaged features & 41.29 & 63.09 & 49.40\\
\;(c) locally-averaged features  & \textbf{44.18} & \textbf{64.89} & \textbf{50.89}\\
\bottomrule
\end{tabular}}}
\vspace{-2pt}
\caption{
\textbf{Comparison of online vocabulary-update approaches.}
The results in the first two sections are with the vanilla version of our method and with the full version in the third section.
}
\vspace{-2pt}
\label{tab:results_vocabulary_update}
\end{table}

\parag{Online vocabulary updates.}
In Tab.\,\ref{tab:results_vocabulary_update}, we compare the approaches for online 
vocabulary updates described \S\,\ref{sec:full_online_training}.
The queue-based solutions achieve in general better results than online k-means. 
Among the queue-based options,
random sampling of locally averaged features, opt.\,(c), provides the best results. 
Its advantage over option (b) with global averaging is more evident with multi-scale BoWs where an extra feature level with a higher resolution and more localized features is used,
in which case global averaging produces visual words that are too coarse.
In all remaining experiments, we use a queue-based vocabulary with option (c).

\parag{Momentum for teacher updates.}
In Table~\ref{tab:results_momentum_coefficient}, we study the sensitivity of our method w.r.t. the momentum $\alpha$ for the teacher updates (Equation \ref{eq:teacher_update}).
We notice a strong drop in performance when decreasing $\alpha$ from $0.9$ to $0.5$ (a rapidly-changing teacher), and to $0$ (the teacher and student have identical parameters), while keeping the initial learning rate fixed ($lr=0.05$).
However, we noticed that this was not due to any cluster/mode collapse issue. 
The issue is that the teacher signal is more noisy at low $\alpha$ because of the rapid change of its parameters. This prevents the student to converge when keeping the learning rate as high as $0.05$. We notice in Table~\ref{tab:results_momentum_coefficient} that a reduction of the learning rate to adapt to the reduction of $\alpha$ reduces the performance gap. This indicates that our method is not as sensitive to the choice of the momentum as MoCo and BYOL were shown to be.

\begin{table}[t!]
\centering
\renewcommand{\figurename}{Table}
\renewcommand{\captionlabelfont}{\bf}
\renewcommand{\captionfont}{\small} 
{\setlength{\extrarowheight}{1pt}\small
{
\begin{tabular}{ l | r | c c | c  }
\toprule
\multicolumn{1}{c|}{}         & & \multicolumn{2}{c|}{Few-shot}                          & \multicolumn{1}{c}{}\\
\multicolumn{1}{c|}{$\alpha$} & \multicolumn{1}{c|}{lr} & \multicolumn{1}{c}{1-shot} & \multicolumn{1}{c|}{1-shot} & \multicolumn{1}{l}{Linear}\\
\midrule
$0.99 \rightarrow 1$ & $0.05$ & \textbf{42.11} & \textbf{62.44} & 45.86\\
\midrule
$0.999$      & $0.05$ & 40.87 & 61.41 & 45.76\\
$0.99$       & $0.05$ & 41.19 & 61.65 & \textbf{46.25}\\
$0.9$        & $0.05$ & 40.79 & 60.92 & 44.89\\
$0.5$        & $0.05$ & 12.70 & 23.20 & 15.41\\
$0.0$        & $0.05$ & 13.19 & 24.85 & 17.47\\
\midrule
$0.5$        & $0.03$ & 39.52 & 60.18 & 43.82\\
$0.0$        & $0.01$ & 33.80 & 55.02 & 39.90\\
\bottomrule
\end{tabular}}}
\vspace{-2pt}
\caption{
\textbf{Influence of the momentum coefficient $\alpha$ used for the teacher updates.}
For these results, we used the vanilla version.
In the ``$0.99 \rightarrow 1$'' row, $\alpha$ is initialized to $0.99$ and annealed to $1.0$ with cosine schedule.
The other entries use constant $\alpha$ values.
}
\vspace{-2pt}
\label{tab:results_momentum_coefficient}
\end{table}

\begin{table}[t!]
\centering
\renewcommand{\figurename}{Table}
\renewcommand{\captionlabelfont}{\bf}
\renewcommand{\captionfont}{\small} 
{\setlength{\extrarowheight}{1pt}\small
{
\begin{tabular}{ c c | r r | r }
\toprule
& & \multicolumn{2}{c|}{Few-shot} & \multicolumn{1}{l}{}\\
Soft & Dyn & \multicolumn{1}{c}{1-shot} & \multicolumn{1}{c|}{5-shot} & \multicolumn{1}{l}{Linear}\\
\midrule
\checkmark & \checkmark & 42.11 & 62.44 & 45.86\\
           & \checkmark & 38.61 & 59.98 & 44.64\\
\checkmark &            &  2.00 &  2.00 &  0.10\\
\bottomrule
\end{tabular}}}
\vspace{-2pt}
\caption{
\textbf{Ablation of dynamic BoW prediction and soft-quantization.}
For these results, we used the vanilla version of our method.
``Soft'': soft assignment instead of hard assignment.
``Dyn'': dynamic weight generation instead of fixed weights.
}
\vspace{-2pt}
\label{tab:results_dyn_soft}
\end{table}

\begin{table}[t!]
\centering
\renewcommand{\figurename}{Table}
\renewcommand{\captionlabelfont}{\bf}
\renewcommand{\captionfont}{\small} 
{\setlength{\extrarowheight}{1pt}\small
{
\begin{tabular}{ l | c | c  }
\toprule
\multicolumn{1}{l|}{Image crops} & Multi-scale &\multicolumn{1}{c}{Linear}\\
\midrule
\;$1 \times 224^2$                        & & 31.39\\
\;$1 \times 224^2$ + cutmix               & & 39.46\\
\;$1 \times 160^2$                        & & 45.86\\
\;$2 \times 160^2$                        & & 47.64\\
\;$5\times 96^2$                          & & 44.24\\
\;$2 \times 160^2$ + $5\times 96^2$       & & 49.64\\
\midrule
\;$2 \times 160^2$                        & \checkmark & 49.00\\
\;$2 \times 160^2$ + $5\times 96^2$       & \checkmark & \textbf{50.89}\\
\bottomrule
\end{tabular}}}
\vspace{-2pt}
\caption{
\textbf{Evaluation of image crop augmentations and of multi-scale BoWs.} See text. 
}
\vspace{-2pt}
\label{tab:results_aug_multibow}
\end{table}

\begin{table}[t!]
\centering
\renewcommand{\figurename}{Table}
\renewcommand{\captionlabelfont}{\bf}
\renewcommand{\captionfont}{\small} 
{\setlength{\extrarowheight}{1pt}\small
{
\begin{tabular}{ l  | r | c   c  | c  }
\toprule
& & \multicolumn{2}{c|}{Few-shot} & \multicolumn{1}{l}{}\\
\multicolumn{1}{l|}{Method} & EP & \multicolumn{1}{c}{$n=1$} & \multicolumn{1}{c|}{$n=5$} & \multicolumn{1}{l}{Linear}\\
\midrule
\;BoWNet                                  & 200 & 33.80 & 55.02 & 41.30\\
\;BoWNet ($160^2$ crops)                  & 200 & 29.26 & 49.68 & 43.59\\
\midrule
\;OBoW (vanilla)     &  80 & 42.11 & 62.44 & 45.86\\
\;OBoW (full)        &  80 & \textbf{44.18} & \textbf{64.89} & \textbf{50.89}\\
\bottomrule
\end{tabular}}}
\vspace{-2pt}
\caption{
\textbf{Comparison with BoW-like methods.}
``EP'': total number of epochs used for pre-training.
Note that the BoWNet method consists of $40$ epochs for teacher pre-training with the RotNet method followed by two BoWNet training rounds of $80$ epochs.
}
\vspace{-2pt}
\label{tab:comparison_sota}
\end{table}

\begin{table*}[t!]
\centering
\renewcommand{\figurename}{Table}
\renewcommand{\captionlabelfont}{\bf}
\renewcommand{\captionfont}{\small} 
{\setlength{\extrarowheight}{1.5pt}\small
{
\begin{tabular}{ l | r | r | c c c | c c c | c c }
\toprule
                            &    &    & \multicolumn{3}{c|}{Linear Classification} & \multicolumn{3}{c|}{VOC Detection} & \multicolumn{2}{c}{Semi-supervised learning}\\
\multicolumn{1}{l|}{Method} & Epochs & Batch & ImageNet &  Places205 & VOC07 & \multicolumn{1}{c}{$\text{AP}^{\text{50}}$} & \multicolumn{1}{c}{$\text{AP}^{75}$} & \multicolumn{1}{c|}{$\text{AP}^{\text{all}}$} & \multicolumn{1}{c}{$1\%$ Labels} & \multicolumn{1}{c}{$10\%$ Labels} \\
\midrule
Supervised                                                      &  100 & 256  & 76.5 & 53.2 & 87.5 & 81.3 & 58.8 & 53.5 & 48.4 & 80.4\\
\midrule
BoWNet~\cite{gidaris2020learning}                                &  325 & 256  & 62.1 & 51.1 & 79.3 & 81.3 & 61.1 & 55.8 & - & -\\
PCL~\cite{li2020prototypical}                                    &  200 & 256  & 67.6 & 50.3 & 85.4 & -    & -    & -       & 75.3 & 85.6\\
MoCo v2~\cite{he2020momentum}                                    &  200 & 256  & 67.5 & -    & -    & 82.4 & 63.6 & 57.0    & - & -\\
SimCLR~\cite{chen2020simple}                                     &  200 & 4096 & 66.8 & -    & -    &  -   & -    & -       & - & -\\
SwAV~\cite{caron2020unsupervised}                                &  200 &  256 & 72.7 & 56.2$^\dagger$ & 87.2$^\dagger$ & 81.8$^\dagger$ & 60.0$^\dagger$ & 54.4$^\dagger$ & 76.7$^\dagger$ & 88.7$^\dagger$\\
BYOL~\cite{grill2020bootstrap}                                   &  300 & 4096 & 72.5 &  -   &  -   &  -   & -    & -       & - & -\\
\midrule
OBoW (Ours)                                                      & 200  & 256  & \textbf{73.8} & \textbf{56.8} & \textbf{89.3} & \textbf{82.9}  & \textbf{64.8} & \textbf{57.9} & \textbf{82.9} & \textbf{90.7}\\
\midrule
PIRL~\cite{misra2020self}                                        &  800 & 1024 & 63.6 & 49.8 & 81.1 & 80.7 & 59.7 & 54.0 & 57.2 & 83.8\\
MoCo v2~\cite{he2020momentum}                                    &  800 & 256  & 71.1 & 52.9 & 87.1 & 82.5 & 64.0 & 57.4 & - & -\\
SimCLR~\cite{chen2020simple}                                     & 1000 & 4096 & 69.3 & 53.3 & 86.4 & -    & -    & -    & 75.5 & 87.8\\
BYOL~\cite{grill2020bootstrap}                                   & 1000 & 4096 & 74.3 &  -   &  -   & -    & -    & -    & 78.4 & 89.0\\
SwAV~\cite{caron2020unsupervised}                                &  800 & 4096 & \textbf{75.3} & 56.5 & 88.9 & 82.6 & 62.7 & 56.1 & 78.5 & 89.9\\
\bottomrule
\end{tabular}}}
\vspace{-0pt}
\caption{
\textbf{Evaluation of ImageNet pre-trained \resnetfifty models}.
The ``Epochs'' and ``Batch'' columns provide the number of pre-training epochs and the batch size of each model respectively.
The first section includes models pre-trained with a similar number of epochs as our model (second section). 
We boldfaced the best results among all sections as well as of only the top two.
For the linear classification tasks, we provide the top-1 accuracy.
For object detection, we fine-tuned Faster R-CNN (R50-C4) on VOC \texttt{trainval07+12} and report detection AP scores by testing on \texttt{test07}.
For semi-supervised learning, we fine-tune the pre-trained models on $1\%$ and $10\%$ of ImageNet and report the top-5 accuracy.
Note that, in this case the ``Supervised'' entry results come from~\cite{zhai2019s} and are obtained by supervised training 
using only $1\%$ or $10\%$ of the labelled data.
All the classification results are computed with single-crop testing. $^\dagger$: results computed by us.
}
\vspace{-0pt}
\label{tab:full_imagenet}
\end{table*}

\parag{Dynamic BoW prediction and soft quantization.}
In Table~\ref{tab:results_dyn_soft}, we study the impact of the dynamic BoW prediction and of using soft assignment for the codes instead of hard assignment. We see that
\textbf{(1)}, as expected, the network is unable to learn useful features without the proposed dynamic BoW prediction, 
i.e., when using fixed weights;
\textbf{(2)} soft assignment indeed provides a performance boost.

\parag{Enforcing context-aware representations.}
In Table~\ref{tab:results_aug_multibow} we study different types of image crops for the BoW reconstruction tasks, as well as the impact of multi-scale BoW targets.
We observe that: \textbf{(1)}
as we discussed in \S\,\ref{sec:contextual_reasoning_skills}, smaller crops that hide significant portions of the original image are better suited for our reconstruction task thus leading to dramatic increase in performance (compare entries $1 \times 224^2$ with the $1 \times 160^2$ and $5 \times 96^2$ entries).
\textbf{(2)}
Randomly sampling two $160 \times 160$-sized crops (entries $2 \times 160^2$) and using $96 \times 96$-sized patches leads to another significant increase in performance.
\textbf{(3)} Finally, employing multi-scale BoWs improves the performance even further.

\parag{BoW-like comparison.}
In Table~\ref{tab:comparison_sota}, we compare our method with the reference BoW-like method BoWNet.
For a fair comparison, we implemented BoWNet both with its proposed augmentations, i.e., using one $224 \times 224$-sized crop with cutmix (``BoWNet'' row), and with the image augmentation we propose in the vanilla version of our method, i.e., one $160 \times 160$-sized crop (``BoWNet ($160^2$ crops)'').
We see that our method, even in its vanilla version, achieves significantly better results, while using at least two times fewer training epochs, which validates the efficiency of our proposed fully-online training methodology.

\subsection{Self-supervised training on ImageNet} \label{sec:large_scale_experiments}

Here we evaluate our method by pre-training with it convnet-based representations on the full ImageNet dataset.
We implement the full solution of our method (as described in \S\,\ref{sec:method_analysis})
using the \resnetfifty (v1)~\cite{he2016deep} architecture.
We evaluate the learned representations on ImageNet, Places205, and VOC07 classification tasks as well as on VOC07+12 detection task and provide results in Table~\ref{tab:full_imagenet}.  
On the ImageNet classification we evaluate on two settings: (1) training linear classifiers with $100\%$ of the data, and (2) fine-tuning the model using $1\%$ or $10\%$ of the data, which is referred to as \emph{semi-supervised} learning.

\parag{Results.}
Pre-training on the full ImageNet and then transferring to downstream tasks is the most popular benchmark for unsupervised representations and thus many methods have configurations specifically tuned on it.
In our case, due to the computational intensive nature of pre-training on ImageNet, no full tuning of 
OBoW took place.
Nevertheless, it achieves very strong empirical results across the board. 
Its classification performance on ImageNet is $73.8\%$, which is substantially better than instance discrimination methods MoCo v2 and SimCLR, and even improves over the recently proposed BYOL and SwAV methods when considering a similar amount of pre-training epochs.
Moreover, in VOC07 classification and Places205 classification, it achieves a new state of the art despite using significantly fewer pre-training epochs than related methods.
On the semi-supervised ImageNet ResNet50 setting, it significantly surpasses the state of the art for $1\%$ labels, and is also better for $10\%$ labels using again much fewer epochs.
On VOC detection, it outperforms previous state-of-the-art methods while demonstrating strong performance improvements over supervised pre-training.

\section{Conclusion}

In this work, we introduce OBoW, a novel unsupervised teacher-student scheme that learns convnet-based representations with a BoW-guided reconstruction task.
By employing an efficient fully-online training strategy and promoting the learning of context-aware representations, it delivers strong results that 
surpass prior state-of-the-art approaches on most evaluation protocols.
For instance, when evaluating the derived unsupervised representations on the Places205 classification, Pascal classification or Pascal object detection tasks, OBoW attains a new state of the art, surpassing prior methods while demonstrating significant improvements over supervised representations.

{\small
\bibliographystyle{ieee_fullname}
\bibliography{egbib}
}

\appendix
\section{Comparing with MoCo for the same image augmentations}

In the main paper we saw that the image augmentation techniques that we designed for our method have a strong positive impact on the quality of the learned representations. 
However, we stress that the performance improvement of our method over state-of-the-art instance-discrimination methods is not simply due a better mix of augmentations. 

For example, in Table~\ref{tab:comparison_mocov2} we compare OBoW with MoCo v2, when the latter is implemented with the same image augmentations as those in the full solution of OBoW.
We see that indeed, although the proposed augmentations also improve MoCo v2, our method is still significantly better, even in its vanilla version that employs simpler augmentations (i.e., only a single $160\times160$-sized crop). 
Therefore, the ability of OBoW to learn state-of-the-art representations is mainly due to its BoW-guided reconstruction formulation.

\section{Visualization of the vocabulary features}

\begin{figure*}
    \centering
    \begin{tabular}{cc}
    \includegraphics[width=0.43\linewidth]{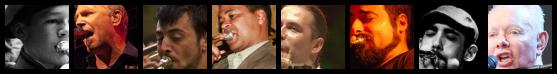} & \includegraphics[width=0.43\linewidth]{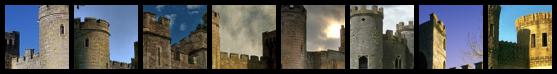} \\
    \includegraphics[width=0.43\linewidth]{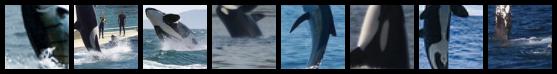} & \includegraphics[width=0.43\linewidth]{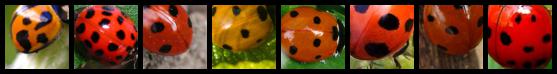} \\  
    \includegraphics[width=0.43\linewidth]{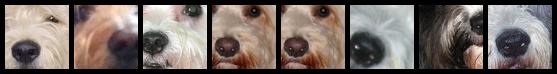} & \includegraphics[width=0.43\linewidth]{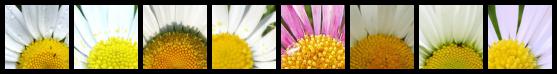} \\
    \includegraphics[width=0.43\linewidth]{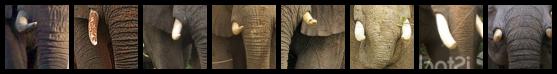} & \includegraphics[width=0.43\linewidth]{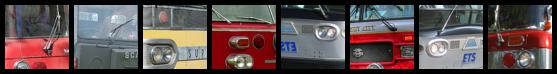} \\  
    \includegraphics[width=0.43\linewidth]{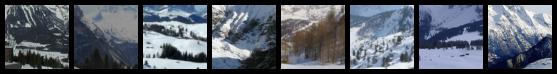} & \includegraphics[width=0.43\linewidth]{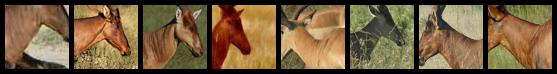} \\ 
    \includegraphics[width=0.43\linewidth]{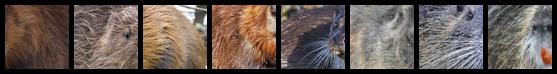} & \includegraphics[width=0.43\linewidth]{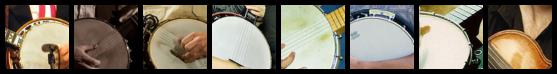} \\ 
    \includegraphics[width=0.43\linewidth]{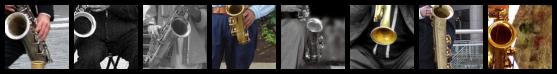} & \includegraphics[width=0.43\linewidth]{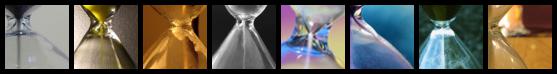} \\  
    \includegraphics[width=0.43\linewidth]{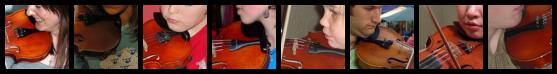} & \includegraphics[width=0.43\linewidth]{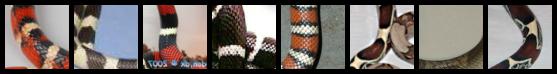} \\
    \includegraphics[width=0.43\linewidth]{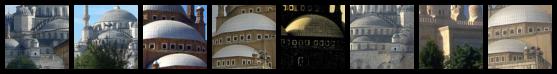} & \includegraphics[width=0.43\linewidth]{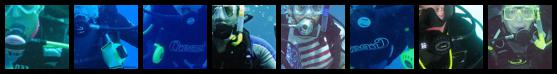} \\
    \includegraphics[width=0.43\linewidth]{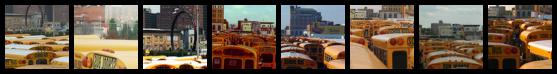} & \includegraphics[width=0.43\linewidth]{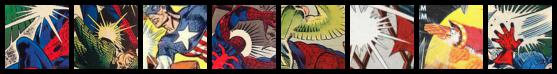} \\    
    \includegraphics[width=0.43\linewidth]{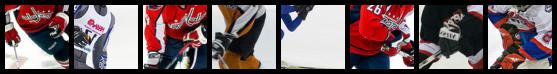} & \includegraphics[width=0.43\linewidth]{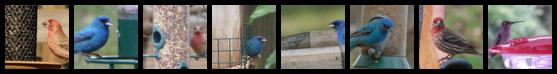} \\      
    \end{tabular}
    \vspace{-4pt}
    \caption{\textbf{Examples of visual-word members from the \texttt{conv5} layer of \resnetfifty.} 
    The visualizations are created by using the state of the queue-based visual-words vocabulary at the end of training. 
    For each visual word, we depict the 8 image patches retrieved from ImageNet with the highest assignment score for that word.
    }
    \label{fig:visual_words_conv5}
    \vspace{-4pt}
\end{figure*}

\begin{figure*}
    \centering
    \begin{tabular}{cc}
    \includegraphics[width=0.43\linewidth]{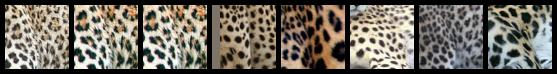} & \includegraphics[width=0.43\linewidth]{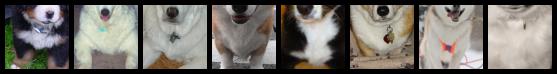} \\
    \includegraphics[width=0.43\linewidth]{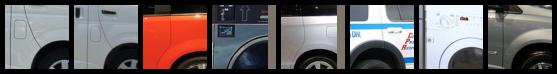} & \includegraphics[width=0.43\linewidth]{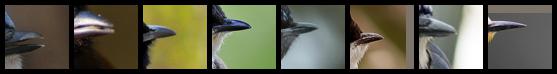} \\  
    \includegraphics[width=0.43\linewidth]{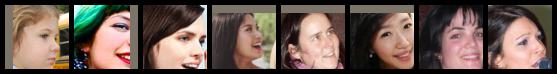} & \includegraphics[width=0.43\linewidth]{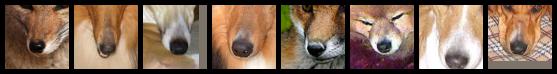} \\
    \includegraphics[width=0.43\linewidth]{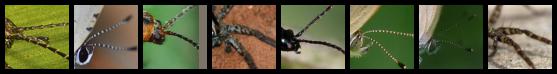} & \includegraphics[width=0.43\linewidth]{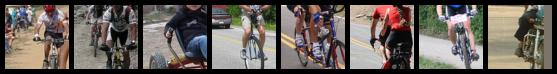} \\  
    \includegraphics[width=0.43\linewidth]{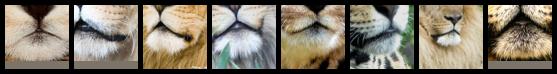} & \includegraphics[width=0.43\linewidth]{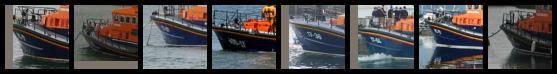} \\ 
    \includegraphics[width=0.43\linewidth]{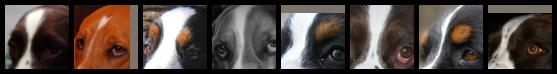} & \includegraphics[width=0.43\linewidth]{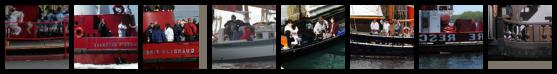} \\ 
    \includegraphics[width=0.43\linewidth]{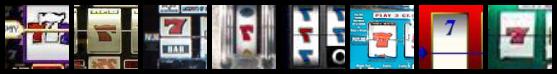} & \includegraphics[width=0.43\linewidth]{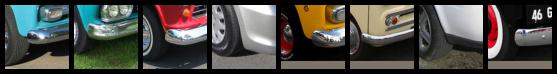} \\  
    \includegraphics[width=0.43\linewidth]{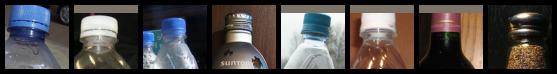} & \includegraphics[width=0.43\linewidth]{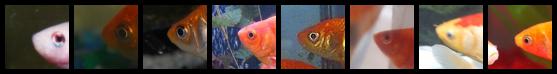} \\
    \includegraphics[width=0.43\linewidth]{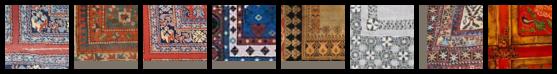} & \includegraphics[width=0.43\linewidth]{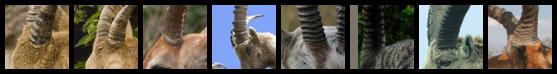} \\
    \includegraphics[width=0.43\linewidth]{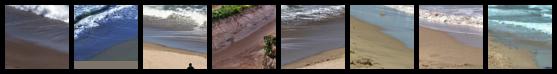} & \includegraphics[width=0.43\linewidth]{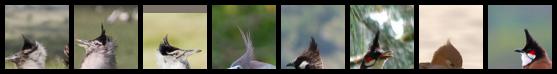} \\
    \includegraphics[width=0.43\linewidth]{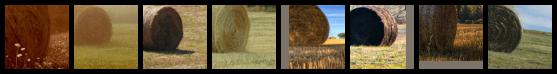} & \includegraphics[width=0.43\linewidth]{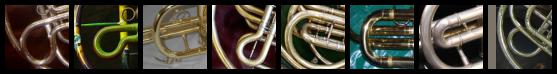} \\
    \end{tabular}
    \vspace{-4pt}
    \caption{\textbf{Examples of visual-word members from the \texttt{conv4} layer of \resnetfifty.} 
    The visualizations are created by using the state of the queue-based visual-words vocabulary at the end of training. 
    For each visual word, we depict the 8 image patches retrieved from ImageNet with the highest assignment score for that word.
    }
    \label{fig:visual_words_conv4}
    \vspace{-4pt}
\end{figure*}

In Figures \ref{fig:visual_words_conv5} and ~\ref{fig:visual_words_conv4} we illustrate visual words from the \texttt{conv5} and \texttt{conv4} teacher feature maps of a \resnetfifty OBoW model trained on ImageNet.
Since we use a queue-based vocabulary that is constantly updated, 
for the visualizations we used the state of the vocabulary at the end of training.
In order to visualize a visual word, we retrieve multiple image patches from images in the ImageNet training set and depict the 8 patches with the highest assignment score for that visual word.
As it can be noticed, visual words encode high level visual concepts.

\begin{table}[t!]
\centering
\renewcommand{\figurename}{Table}
\renewcommand{\captionlabelfont}{\bf}
\renewcommand{\captionfont}{\small} 
{\setlength{\extrarowheight}{1.5pt}\small
{
\begin{tabular}{ l  | r | c   c  | c  }
\toprule
& & \multicolumn{2}{c|}{Few-shot} & \multicolumn{1}{l}{}\\
\multicolumn{1}{l|}{Method} & EP & \multicolumn{1}{c}{$n=1$} & \multicolumn{1}{c|}{$n=5$} & \multicolumn{1}{l}{Linear}\\
\midrule
\;OBoW (vanilla)     &  80 & 42.11 & 62.44 & 45.86\\
\;OBoW (full)        &  80 & \textbf{44.18} & \textbf{64.89} & \textbf{50.89}\\
\midrule
\;MoCo v2            & 80 & 24.75 & 43.89 & 35.00 \\
\;MoCo v2 (our aug.) & 80 & 36.90 & 55.87 & 43.13 \\
\bottomrule
\end{tabular}}}
\vspace{-2pt}
\caption{
\textbf{Comparison with MoCo v2 for the same image augmentations.}
``EP'': total number of epochs used for pre-training.
The results are obtained by training ResNet18-based models on $20\%$ of ImageNet, similar to Sections 4.1 and 4.2 of the main paper.
``MoCo v2 (our aug.) '' is a MoCo v2 model implemented with the same augmentations that we use in the full version of our work, i.e., two $160 \times 160$-sized crops plus five $96 \times 96$-sized patches.
}
\vspace{-2pt}
\label{tab:comparison_mocov2}
\end{table}

\section{Implementation details} \label{sec:appendix_implementation_details}

\subsection{Image augmentation during pre-training}

In Section~\ref{sec:contextual_reasoning_skills} of the main paper,
we described the two types of image crops that we extract from a training image in order to train the student network with them.
In addition, beyond image cropping, similar to SimCLR~\cite{chen2020simple}, we also applied color jittering, color-to-grayscale conversion, Gaussian blurring and horizontal flipping as augmentation techniques. 
All implementation details are provided in Section \ref{sec:aug_pytorch} in the form of PyTorch code.

\subsection{Evaluation protocols in Section 4.1}

To evaluate the quality of the learned representations, we use two protocols.
\textbf{(1)} The first protocol consists in freezing the convnet and then training on its features $1000$-way linear classifiers for the ImageNet classification task.
The classifier is applied on top of the $512$-dimensional feature vectors produced from the final global pooling layer of ResNet18.
It is trained with SGD for $50$ epochs using a learning rate of $10$ that is divided by a factor of $10$ every $15$ epochs.
The batch size is $256$ and the weight decay $2\rm e{-}6$. 
For fast experimentation, we train the linear classifier with precached features extracted from the $224 \times 224$ central crop of the image and its horizontally flipped version. 
\textbf{(2)}
For the second protocol, we use a few-shot episodic setting~\cite{vinyals2016matching}. 
We choose $300$ classes from ImageNet and run $200$ episodes of $50$-way few-shot classification tasks.
Essentially, for each episode, we randomly select $50$ classes from the $300$ ones and, for each of these selected classes, $n$ training examples and $m=1$ test example (both randomly sampled from the validation images of ImageNet). 
For $n$, we use $1$ and $5$ examples corresponding to $1$-shot and $5$-shot classification settings, respectively.
The $m$ test examples per class are classified using a cosine-distance Prototypical-Networks~\cite{snell2017prototypical} classifier applied on top of the frozen self-supervised representations. We report the mean accuracy over the $200$ episodes.
The purpose of this metric is to analyze the ability of the representations to be used for learning with few training examples.
Furthermore, it has the advantage of not requiring tuning of any hyper-parameters, such as the learning rate of a linear classifier, the number of training steps, \etc

\subsection{Self-supervised training on ImageNet}

Here we provide implementation details for the pre-training of the \resnetfifty-based OBoW model that we use in Section~\ref{sec:large_scale_experiments} of the main paper.
We present the full implementation of our method, which includes multi-scale BoWs from the \texttt{conv4} and \texttt{conv5} layers of \resnetfifty, and extraction of two crops of size $160 \times 160$ plus five
patches of size $96 \times 96$ per training image.
To extract BoW targets, 
we use $K=8192$ as vocabulary size and we ignore the local feature vectors on the edge / border of the teacher's feature maps. 
The momentum coefficient $\alpha$ for the teacher updates is initialized at $0.99$ and is annealed to $1.0$ during training with a cosine schedule.
Finally, the hyper-parameters $\kappa$ and $\delta_{\mathrm{base}}$ are set to $8$ and $1/15$ respectively.

We train the model for $200$ training epochs with SGD using $1\rm e{-}4$
weight decay and $256$-sized mini-batches.
As a learning-rate schedule, we warm up the learning rate from $0$ to $0.03$ with linear annealing during the first $10$ epochs and then, for the remaining $190$ epochs, we decrease it from $0.03$ to $0.00003$ with cosine-based schedule.  
To train the model we use 4 Tesla V100 GPUs with data-distributed training (i.e., the mini-batch is divided across the 4 GPUs) while keeping the batch-norm statistics synchronized across all GPUs (i.e., use the \texttt{SyncBatchNorm} units of PyTorch).

\subsection{Evaluation protocols in Section 4.3}

Here we describe the evaluation protocols that we use in Section~\ref{sec:large_scale_experiments} of the main paper.

\parag{ImageNet linear classification.}
In this case, 
we evaluate the performance on the $1000$-way ImageNet classification task by applying a linear classifier on top of the $2048$-dimensional frozen features of the \texttt{pool5} layer of \resnetfifty.
We train the linear classifier using SGD for $100$ training epochs with $0.9$ momentum, $0$ weight decay, $1024$-sized mini-batches and cosine learning-rate schedule initialized at $10.0$.
We use the typical image augmentations used for the fully-supervised training of \resnetfifty models on this dataset.

\parag{Places205 linear classification.}
For this protocol, 
we evaluate the performance on the $205$-way Places classification task by applying a linear classifier on top of the $2048$-dimensional frozen features of the \texttt{pool5} layer of \resnetfifty.
We follow the guidelines of \cite{goyal2019scaling} and train the linear classifier using SGD for $14$ training epochs and a learning rate of $0.01$ that is multiplied by $0.1$ after $5$ and $10$ epochs.
The batch size is $256$ and the weight decay is $1\rm e{-}4$.

\parag{VOC07 linear classification with SVMs.}
Here we evaluate on the VOC07 classification task by training and testing linear SVMs on top of the $2048$-dimensional frozen features of the \texttt{pool5} layer.
To this end, we use the publicly available code for benchmarking self-supervised methods provided in \cite{goyal2019scaling} that trains the SVMs using the VOC07 train+val splits and tests them using the VOC07 test split. 

\parag{Semi-supervised learning setting on ImageNet.}
For this semi-supervised setting, we fine-tune the self-supervised \resnetfifty model (pre-trained on all ImageNet unlabelled images) on $1\%$ or $10\%$ of ImageNet labelled images. 
We use the same $1\%$ and $10\%$ splits as in SimCLR (i.e., we downloaded and use the split files of their official code release).
We train using SGD with $256$-sized mini-batches, $0$ weight decay, and two distinct learning rates for the classification head and the feature extractor trunk network components respectively. 
Specifically, in the $1\%$ setting, we use $40$ epochs and the initial learning rates $0.5$ and $0.0002$ for the classification head and feature extractor trunk components, respectively, which are then multiplied by a factor of $0.2$ after $24$ and $32$ epochs.
In the $10\%$ setting, we use $20$ epochs and the initial learning rates $0.5$ and $0.0002$ for the classification head and feature extractor trunk components, respectively, which are multiplied by a factor of $0.2$ after $12$ and $16$ epochs.

\parag{VOC object detection.}
Here we evaluate the utility of OBoW on a complex downstream task: 
object detection. We follow the setup considered in prior works~\cite{caron2020unsupervised, gidaris2020learning, goyal2019scaling, he2020momentum, misra2020self}: we fine-tune the pre-trained OBoW with a Faster R-CNN~\cite{ren2015faster} model using a \resnetfifty  backbone~\cite{he2017mask} (R50-C4 in Detectron2~\cite{wu2019detectron2}). We use
the fine-tuning protocol and most hyper-parameters from He~\etal~\cite{he2020momentum}: fine-tune on \texttt{trainval07+12} and evaluate on \texttt{test07}. In detail, we train with mini-batches of size $16$ across $4$ GPUs for $24$K steps, using \texttt{SyncBatchNorm} to finetune BatchNorm parameters, as well as inserting an additional BatchNorm layer for the RoI head after  \texttt{conv5}, i.e., \texttt{Res5ROIHeadsExtraNorm} layer in Detectron2. The initial learning rate $0.01$ is warmed-up with a slope of $1\rm e{-}3$ for $100$ steps and then reduced by a factor of $10$ after $18$K and $22$K steps. We report results for the final checkpoint averaged over 3 different runs. 

\section{On-line k-means vocabulary updates} \label{sec:appendix_online_kmeans}

As explained in the main paper, one of the explored choices for updating the vocabulary is to use online k-means after each training step.
Specifically, as proposed in VQ-VAE~\cite{oord2017neural, razavi2019generating}, 
we use exponential moving average for vocabulary updates.
In this case, for each mini-batch, we compute the number $n_k$ of feature vectors assigned to each cluster $k$ and $\mathbf{m}_k$ the element-wise sum of all feature vectors assigned to cluster $k$ and update
\begin{align}
N_k & \gets \gamma N_k  + (1 - \gamma) n_k, \\
\mathbf{M}_k & \gets \gamma \mathbf{M}_k  + (1 - \gamma) \mathbf{m}_k,
\label{eq:kmeans_updates}
\end{align}
with $\gamma = 0.99$. The $k^{\rm th}$ visual word of the vocabulary $V$ satisfies $\vv_k = {\mathbf{M}_k}/{N_k}$.
A critical issue that arises in this case is that, as training progresses, the features distribution changes over time. The visual words computed by online k-means do not adapt to this distribution shift leading to extremely unbalanced cluster assignments and even to assignments that collapse to a single cluster. In order to counter this effect, we investigate two different strategies: 

\parag{(a) Detection and replacement of rare visual words.}
In this case, for each visual word we keep track of the time of its most recent assignment as closest cluster centroid to a feature vector.
If more than $1000$ training steps have passed since then, then we replace it with a local feature vector randomly sampled with uniform distribution from the current mini-batch.

\parag{(b) Enforcing uniform assignments via Sinkhorn optimization.}
Let $\vx_1, \ldots, \vx_b$ be the $b$ images of the current mini-batch and $D$ be the $K \times B$ matrix ($B = h_{\ell} \times w_{\ell} \times b$) whose $i^{\rm th}$ row $D_i$ contains the squared distances between all the local features in the mini-batch (across all images and spatial dimensions) and the $i^{\rm th}$ visual word: $D_i = \big[\, \| \mathrm{T}(\vx_1)[1] - \mathbf{v}_i \|_2^2, \ldots, \| \mathrm{T}(\vx_b)[h_{\ell} \times w_{\ell}] - \mathbf{v}_i \|_2^2,\big]$. 
Similarly to \cite{asano2019selflabelling,caron2020unsupervised}, 
we compute the assignment codes by solving the regularised optimal transport problem
$\min_{Q \in \mathcal{Q}} \sum_{i, j} Q_{i, j} D_{i, j} + \varepsilon \, Q_{i, j} \log Q_{i, j}$,
where $\varepsilon$ is a coefficient that controls the softness of the assignments. 
The set $\mathcal{Q}$ permits us to enforce uniform assignments among all the visual words and satisfies
$\mathcal{Q} = \{Q \in \mathbb{R}_{+}^{K \times B} | Q \mathbf{1}_{B} = \frac{1}{K} \mathbf{1}_{K}, Q^{\top}\mathbf{1}_K = \frac{1}{B} \mathbf{1}_B\}$,
where $\mathbf{1}_{K}$ and $\mathbf{1}_{B}$ are vectors of length $K$ and $B$, respectively, with all entries equal to $1$. 
We compute $Q$ with the Sinkhorn algorithm~\cite{cuturi2013sinkhorn} and use the resulting assignment codes for the on-line k-means updates and for the computation of the BoW targets.

\begin{table}[b!]
\centering
\renewcommand{\figurename}{Table}
\renewcommand{\captionlabelfont}{\bf}
\renewcommand{\captionfont}{\small} 
{\setlength{\extrarowheight}{1.5pt}\small
{
\begin{tabular}{ l | c c c c }
\toprule
& \multicolumn{1}{r}{Sup.} & \multicolumn{1}{r}{OBoW} & \multicolumn{1}{r}{MoCo v2} & \multicolumn{1}{r}{BYOL}\\
\midrule
\;Epochs           & 100 & 200 & 800 & 300\\
\midrule
\multicolumn{5}{l}{\textbf{Measured with $256$-sized mini-batches}} \\
\;Time per epoch     & 1.00 & 3.91 &  1.58 &  3.47\\
\;Training time      & 1.00 & 7.82 & 12.64 & 10.41\\
\;Memory per GPU     & 1.00 & 2.00 &  1.13 &  1.72\\
\midrule
\multicolumn{5}{l}{\textbf{ImageNet linear classification accuracy}} \\
\;batch size = $256$  & \multicolumn{1}{c}{76.5} & \multicolumn{1}{c}{73.8} & \multicolumn{1}{c}{71.1}  & \multicolumn{1}{c}{-}\\
\;batch size = $4096$ & \multicolumn{1}{c}{-} & \multicolumn{1}{c}{-} &  \multicolumn{1}{c}{-} & \multicolumn{1}{c}{72.5$^\dagger$}\\
\bottomrule
\end{tabular}}}
\vspace{-0pt}
\caption{\textbf{Time and memory consumption relative to supervised training.}
``Sup.'' is the supervised ImageNet training.
To measure the time and memory consumption, for all methods we used \resnetfifty-based implementations, $256$-sized mini-batches and data-distributed training with 4 Tesla V100 GPUs.
We measured the time consumption based on a single training epoch (``Time per epoch''). 
We also provide the projected time for the full training of a method (``Training time''), which is estimated based on the specified number of training epochs (``Epochs''). 
For OBoW, we used its full implementation. 
$^\dagger$: for BYOL we provide the time and memory consumption w.r.t. $256$-sized mini-batches, but BYOL uses $4096$-sized mini-batches to achieve the reported ImageNet classification accuracy.
So, in reality BYOL has higher total GPU memory requirements.
}
\vspace{-0pt}
\label{tab:complexity_comparison}
\end{table}

\section{Time and memory consumption}

In Table~\ref{tab:complexity_comparison}, we provide the time and memory consumption of our method as well as of MoCo v2 and BYOL.
We observe that OBoW achieves state-of-the-art results in less time (``Training time'' row) than the competing methods.
In terms of GPU memory consumption, with $256$-sized mini-batches our method requires $15775$Mb per GPU in a 4-GPU machine (or $8901$Mb per GPU in a 8-GPU machine).

\section{COCO detection and instance segmentation}

In Table~\ref{tab:coco_results} we evaluate the learned representations on the downstream tasks of object detection and instance segmentation on COCO~\cite{lin2014microsoft}.
To this end, we fine-tune the pre-trained representations with a Mask R-CNN~\cite{he2017mask} model with a \resnetfifty backbone and feature pyramid networks~\cite{lin2017feature} (Mask R-CNN R50-FPN) implemented in Detectron2.
We train the Mask R-CNN model on \texttt{train2017} for $12$ epochs ($1\times$ schedule) and report the box detection AP (\apbbox{~}) and instance segmentation AP (\apmask{~}) on COCO \texttt{val2017}.
Similar to VOC detection experiments, BatchNorm layers are fine-tuned and synchronized.
We see that OBoW achieves better or comparable results to prior methods despite the fact that it used only 200 pre-training epochs. 

\begin{table}[h!]
\centering
\renewcommand{\figurename}{Table}
\renewcommand{\captionlabelfont}{\bf}
\renewcommand{\captionfont}{\small} 
{\setlength{\extrarowheight}{1.5pt}\small
{
\begin{tabular}{ l | r | r | c c }
\toprule
method & Epochs & Batch & \apbbox{~} & \apmask{~}\\
\midrule
Supervised & 100 & 256 & 39.6 & 35.6\\
VADeR \cite{pinheiro2020unsupervised} & 600 & 128 & 39.2 & 35.6\\ 
SimCLR \cite{chen2020simple}         & 1000 & 4096 & 39.7 & 35.8\\ 
MoCo v2	\cite{chen2020improved}     &  800 &  256 & 40.4 & 36.4\\
InfoMin \cite{tian2020makes}    &  200 &  256 & 40.6 & 36.7\\ 
BYOL \cite{grill2020bootstrap}    & 1000 & 4096 & 41.6 & 37.2\\
SwAV \cite{caron2020unsupervised} &  800 & 4096 & 41.6 & 37.8\\ 
\midrule
OBoW (ours) & 200 & 256 & 40.8 & 36.4\\
\bottomrule
\end{tabular}}}
\vspace{-0pt}
\caption{\textbf{Results of object detection and instance segmentation on COCO.} 
We report the box detection AP (\apbbox{~}) and instance segmentation AP (\apmask{~}) on \texttt{val2017}.
All methods are pretrained on ImageNet and then fined-tuned on COCO for $12$ epochs ($1\times$ schedule).
We use the \resnetfifty-based Mask-RCNN model equipped with feature pyramid networks (Mask R-CNN R50-FPN). 
The ``Epochs'' and ``Batch'' columns provide the number of pre-training epochs and the batch size of each model respectively.
}
\vspace{-0pt}
\label{tab:coco_results}
\end{table}

\onecolumn

\clearpage
\onecolumn
\section{PyTorch code of Image augmentations} \label{sec:aug_pytorch}
Here we provide the PyTorch implementation of the image augmentations used in our work.
\begin{lstlisting}[language=Python]
import random
import torch
import torchvision.transforms as T
from PIL import ImageFilter

class CropImagePatches:
    """Crops from an image 3 x 3 overlapping patches."""
    def __init__(self, patch_size=96, patch_jitter=24, num_patches=5):
        self.patch_size = patch_size
        self.patch_jitter = patch_jitter
        self.num_patches = num_patches

    def __call__(self, img):
        _, height, width = img.size()
        split_per_side = 3
        offset_y = (height - self.patch_size - self.patch_jitter) // (split_per_side-1)
        offset_x = (width - self.patch_size - self.patch_jitter) // (split_per_side-1)

        patches = []
        for i in range(split_per_side):
            for j in range(split_per_side):
                y_top = i * offset_y + random.randint(0, self.patch_jitter)
                x_left = j * offset_x + random.randint(0, self.patch_jitter)
                y_bottom = y_top + self.patch_size
                x_right = x_left + self.patch_size
                patches.append(img[:, y_top:y_bottom, x_left:x_right])

        if self.num_patches < (split_per_side * split_per_side):
            indices = torch.randperm(len(patches))[:self.num_patches]
            patches = [patches[i] for i in indices.tolist()]

        return torch.stack(patches, dim=0)

class StackMultipleViews:
    def __init__(self, transform, num_views):
        self.transform = transform
        self.num_views = num_views

    def __call__(self, img):
        return torch.stack([self.transform(img) for _ in range(self.num_views)], dim=0)

class GaussianBlur:
    def __init__(self, sigma=[.1, 2.]):
        self.sigma = sigma

    def __call__(self, img):
        sigma = random.uniform(self.sigma[0], self.sigma[1])
        return img.filter(ImageFilter.GaussianBlur(radius=sigma))

normalize = T.Normalize(mean=[0.485, 0.456, 0.406], std=[0.229, 0.224, 0.225])

# Define the transformations for extracting the central from of the original image.
transform_original_image = T.Compose([
    T.Resize(256),
    T.CenterCrop(224),
    T.RandomHorizontalFlip(),
    T.ToTensor(),
    normalize])

# Define the transformations for generating two 160x160-sized image crops.
transform_two_160x160_image_crops = StackMultipleViews(
    transform=T.Compose([
        T.RandomResizedCrop(160, scale=[0.08, 0.6]),
        T.RandomApply([T.ColorJitter(0.4, 0.4, 0.4, 0.1)], p=0.8),
        T.RandomGrayscale(p=0.2),
        T.RandomApply([GaussianBlur(sigma=[0.1, 2.0])], p=0.5),
        T.RandomHorizontalFlip(),
        T.ToTensor(),
        normalize]),
    num_views=2)

# Define the transformations for generating two 160x160-sized image crops.
transform_five_96x96_image_patches = T.Compose([
    T.RandomResizedCrop(256, scale=[0.6, 1.0]),
    T.RandomApply([T.ColorJitter(0.4, 0.4, 0.4, 0.1)], p=0.8),
    T.RandomGrayscale(p=0.2),
    T.RandomHorizontalFlip(),
    T.ToTensor(),
    normalize,
    CropImagePatches(patch_size=96, patch_jitter=24, num_patches=5)])
\end{lstlisting}

\end{document}